\documentclass{article}
\usepackage{amsmath,amsfonts}
\usepackage{algorithmic}
\usepackage{algorithm}
\usepackage{array}
\usepackage[caption=false,font=normalsize,labelfont=sf,textfont=sf]{subfig}
\usepackage{textcomp}
\usepackage{stfloats}
\usepackage{url}
\usepackage{verbatim}
\usepackage{graphicx}
\usepackage{cite}

\usepackage{xspace}

\newtheorem{definition}{Definition}

\usepackage[dvipsnames]{xcolor}
\usepackage{todonotes}

\usepackage{listings}
\begin{document}

\title{Improving Noise Robustness through Abstractions and its Impact on Machine Learning}

\author{Alfredo Ibias$^{1}$ \and Karol Capa{\l}a$^{1}$  \and Varun Ravi Varma$^1$ \and Anna Dro{\.z}d{\.z}$^1$  \and Jose Sousa$^1$}

\date{
    $^1$Personal Health Data Science, Sano - Centre for Computational Personalised Medicine \\ \texttt{j.sousa@sanoscience.org}\\[2ex]}%

\maketitle

\begin{abstract}
Noise is a fundamental problem in learning theory with huge effects in the application of Machine Learning (ML) methods, due to real world data tendency to be noisy. 
Additionally, introduction of malicious noise can make ML methods fail critically, as is the case with adversarial attacks. 
Thus, finding and developing alternatives to improve robustness to noise is a fundamental problem in ML. 
In this paper, we propose a method to deal with noise: mitigating its effect through the use of data abstractions. 
The goal is to reduce the effect of noise over the model's performance through the loss of information produced by the abstraction. 
However, this information loss comes with a cost: it can result in an accuracy reduction due to the missing information. 
First, we explored multiple methodologies to create abstractions, using the training dataset, for the specific case of numerical data and binary classification tasks. 
We also tested how these abstractions can affect robustness to noise with several experiments that explore the robustness of an Artificial Neural Network to noise when trained using raw data \emph{vs} when trained using abstracted data. 
The results clearly show that using abstractions is a viable approach for developing noise robust ML methods.
\end{abstract}


\section{Introduction}
Noisy data is a common problem affecting statistical analysis of data across all fields of science.
The domain of signals and systems defines noise as distortion or corruption of data that leads to false conclusions, and it can present itself in two ways: as deterministic or as random noise.
Deterministic noise is usually easily dealt with as bias in the dataset.
Random noise is usually difficult to detect and account for.

This problem extends to the domain of Machine Learning (ML) and affects the inference capabilities of ML models, causing models to generalise poorly on the task they are designed and trained for.
The worst case of this phenomenon are adversarial attacks, which could trick a ML model to misclassify a data point by adding a noise mask that makes small changes to the original values~\cite{khamaiseh2022adversarial}.
Thus, dealing with noise presents itself as a fundamental task for the wider application of ML methods, as creating curated datasets and data cleaning processes is not always possible or feasible.

An example of a noise prone dataset is the measurements performed using wearable devices for healthcare purposes.
In this case the noise is introduced by the fact that wearable devices need a very specific configuration to work as intended, but people, usually, are not able to stay in such configuration for a long time.
Therefore, many of the measures performed by these devices are performed in sub-optimal conditions, which leads to the introduction of noise and errors in the recorded data~\cite{cosoli2021wearable}.
This is not the only example of noisy data in the domain of healthcare, neither in other fields, but it clearly shows how noise can be unavoidable in some settings.

In ML, abstracted data is seen as low quality information which hinders the learning process, resulting in sub-optimal performance at the task the model is trained for.
Therefore its use is currently limited to encoding non-numeric data, such as labels and text.
In this work, we want to extend the use of abstractions to numerical data and explore its influence on the robustness to noise and reduction in accuracy when used in tandem with ML models.
More precisely, we propose to use abstractions as a way to make ML models robust to noise.
To this end, we explore the trade-off between accuracy and noise robustness that could be fundamental for certain ML applications.
We expect data abstraction will help mitigate the effect of noise present in data, thus allowing for better generalisation on the training dataset.
However, we are aware of the downside of information loss through the abstraction process.
This could be (and has been, as far as we know) a huge deterrent to the application of abstractions as a preprocessing step in ML models.

We decided to choose abstractions for their following qualities.
First, they generalise the input data to a higher order representation, which helps clean impurities (that is, noise) from the data.
Second, they allow for abstract reasoning, as they do not depend on low level data values.
This could improve interpretability, as abstract concepts are easily understandable.
Third, they simplify problems and computations, as they discretise a potentially infinite n-dimensional space into a finite space of the same dimension.
Finally, they also help to generate datasets compliant with the General Data Regulation Protection (GDPR) allowing for their use without the risk of data leakages.
This is done by creating a map of the raw data that cannot be reverted, thus essentially anonimysing the original data.
This last bit is actually crucial in some ML application areas, as healthcare, were raw data is not easily accessible or shareable.

In this paper, we first explore which option is the best one to generate abstractions.
Abstractions can be generated in multiple ways, and each one of them keep different characteristics of the original data.
Thus, it is useful to first explore which kind of abstractions would provide better results when used with an ML method, from an accuracy perspective.
Hence, we explored the use of static binning, quantiles, ROC Curves and K-means clustering to generate abstractions, and our experiments determined that ROC Curves and quantiles were the best options for generating abstractions, depending on the dataset.

After this initial experimentation, we focused on exploring the effect of abstractions on the performance of an Artificial Neural Network (ANN).
ANNs are derived from a sub-field of Artificial Intelligence (AI) research which aims to simulate intelligent behaviour by mimicking biological NNs.
In other words, ANNs focus on reproducing ``how we do'' instead of ``what we do''.
Due to this approach they are heavily reliant on the volume of data available, that is, they are data driven.
This data driven approach aims to reduce the uncertainty from the mimicking process, whereas noise in data increases the uncertainty.
ANNs have been used with different levels of success for various tasks, and, are probably, the most used ML approach~\cite{prieto2013advances}.

In our work we consider the scenario where we have a dataset wherein we are unsure of the presence of noise, and we want to develop a ML method for binary classification.
This dataset is composed of purely numerical features, without any categorical or text feature.
In this scenario, we assume that the class labels are not affected by the noise, but any other feature is.
We also assume that this noise can be present everywhere, but that its amplitude is not very high, at most $10\%$ of the range of the data.

We performed several experiments to evaluate our proposal, allowing for comparisons in performance of ML models between raw, abstracted and noisy data.
The results of such experiments show that the loss in accuracy while using abstractions is almost negligible.
At the same time, a noticeable improvement in noise robustness is observed. 
Moreover, the evolution of the abstraction proposal for different scenarios is discussed.
Specifically, the behaviour of ANNs across three scenarios is explored: when the noise in the training dataset is reproduced in the test dataset, when the noise in the training dataset is not reproduced in the test dataset, and when a very well curated and clean dataset is used for training but the test dataset is noisy.
In all the cases, we observe that using an abstraction is better than using the raw data. 
This is due to the abstraction removing noise-related information, therefore enabling the model to generalise in a better way than when using raw data.

The subsequent sections of the paper present some related work in Section~\ref{sec:relwk} followed by reintroducing fundamental knowledge in Section~\ref{sec:prel}.
Section~\ref{sec:prop} introduces the details of our proposal, continuing with the summaries of our experiments in Section~\ref{sec:exps}.
In Section~\ref{sec:tstv} the threats to the validity of our experiments are raised.
Finally, Section~\ref{sec:conc} presents the conclusions of our work.

\section{Related Work}\label{sec:relwk}
Traditional approaches to dealing with noise involve cleaning the noise from the dataset.
The simplest methods involve identifying and subtracting noisy data from the dataset.
Tools such as \emph{yonder}~\cite{chen2022yonder} and \emph{Denoising Auto-Encoders}~\cite{zhao2020semisupervised}, make assumptions about the structure of noise in the data in order to construct a function enabling noise removal.
However, these methods fail to remove noise that does not follows their base assumption.

In order to achieve better understanding of the data, services like Amazon Mechanical Turks~\cite{paolacci2014current, wood2021using} and Google CrowdSource~\cite{google_crowdsource} give users access to massive human workforce working remotely to classify the data and provide relatively cleaner input for modelling.
The reliability of the data curated from such sources is questionable, since there is a high chance of human error in the curating process~\cite{ wazny2017crowdsourcing}.
Therefore, they fail to provide noise free datasets.

Despite all these efforts, noise continues to be a big problem in the field of data-centric learning.
Until date, the best approach to deal with noisy data is training models incorporating such noise in the data~\cite{Adeli2019Semi}.
Alternative methods have been developed too, but they are less explored. For example, there are proposals like using noise measures to perform adversarial regularization~\cite{Fatras2022Wasserstein}, filtering noisy samples from the dataset~\cite{Unal2023Learning}, or using matrix decomposition methods to model noise~\cite{Zhang2021Bayesian}.
However, so far abstractions have been used only as a method to reduce dimensionality~\cite{ougiaroglou2012simple}, where the proposal was to use the k-means clustering algorithm as a data reduction technique to improve the k-Nearest Neighbour classification algorithm performance.
Thus, to the best of our knowledge, no one has tried to deal with noise through abstractions.

\section{Theoretical Background}\label{sec:prel}
In this section, we revisit some basic Machine Learning (ML) methods that we are going to use in our experiments to test how abstracted data affects their performance.
The ML methods we use in our experiments are Logistic Regression, Random Forest, Support Vector Machines (SVM), and Artificial Neural Networks (ANNs).

Additionally, we need to introduce the measure we are going to use to test how well the different ML methods perform on new, previously unseen inputs. 
To that end, a set of inputs and their corresponding classes are selected, and with this validation set, we compute the \emph{classification accuracy} of the method. Specifically, we compute the balanced accuracy.
Since we apply the ML method to a binary classification task between \textit{positive} and \textit{negative} result classes, the balanced accuracy can be expressed in the classical form.
\begin{equation}
    \label{eqn:BA}
BA = \frac{\frac{TP}{TP+FN}+\frac{TN}{TN+FP}}{2}
\end{equation}
where 
\begin{itemize}
    \item True positives (TP): number of correctly classified samples of the positive class.
    \item False positives (FP): number of incorrectly classified samples of the positive class.
    \item True negatives (TN): number of correctly classified samples of the negative class.
    \item False negatives (FN): number of incorrectly classified samples of the negative class.
\end{itemize}
Equation (\ref{eqn:BA}) can be understood as an average of the probabilities that the positive and negative predictions were  correct.
For the balanced dataset, i.e., containing the same number of elements from positive and negative classes, formula for balanced accuracy reduces to the classical formula of accuracy
\begin{equation}\label{eq:ROCaccuracy}
    ACC = \frac{TP+TN}{TP+TN+FP+FN},
\end{equation}

\subsubsection{Logistic Regression}
Logistic Regression~\cite{walker1967estimation} is a classification method that models a binary output variable as a logistic function of the following form
$$\displaystyle \frac{1}{1+e^{-x}}$$
This function is limited to the range $(0,1)$.
Logistic Regression is trained using Maximum Likelihood Estimation (MLE)~\cite{pan2002maximum} as loss function.
We choose Logistic Regression as one of our comparison methods because it was one of the first ML methods to be developed, and it allows for a very basic first comparison.

\subsubsection{Random Forest}
Random Forests~\cite{breiman2001random} is one of the simplest, Decision Tree based ensemble learning methods for classification.
During training, a multitude of trees are constructed, and the class label is decided on the basis of the votes of each tree.
Decision Trees use the concept of entropy~\cite{maszczyk2008comparison} to create splits or decision boundaries for each feature in the data; the choice of feature is dependent on the Maximal Entropy Principle~\cite{guiasu1985principle}.
We choose Random Forest as one of our comparison methods because it is widely used in industry (owing to its explainable decision making nature) and because it performs decently well on average.

\subsubsection{Support Vector Machines}
The Support Vector Machine (SVM)~\cite{cortes1995support} maps training examples to points in a hyperspace to find the division boundary that maximises the gap between two (or more) categories.
This is done through ``kernel techniques'' (also known as \emph{the kernel trick}) that allows the method to work in a higher order space without needing to actually transform the points to such space.
Due to this fact, SVMs are very dependent on the kernel you choose for them, as different kernels allow for different shapes of the division boundary.

\subsubsection{Artificial Neural Networks}
Artificial Neural Networks (ANNs)~\cite{goodfellow2016deep} are a typical ML algorithm that consists of a set of interconnected neurons, deployed in a layered architecture, working together to perform classification (or regression) tasks.
These neurons are simple agents that perform a weighted sum of the outputs of the neurons of the previous layer, and then apply a non-linear function, called an activation function, to the result.
Usually, all neurons in a layer use the same activation function.
The layers of a neural network are classified into $3$ types:
\begin{itemize}
    \item Input layer: for receiving the input.
    \item Hidden layers: to process the data.
    \item Output layer: to perform the final classification/regression operation.
\end{itemize}   
To train the layers of neurons, ANNs use the \emph{back-propagation}~\cite{rumelhart1986learning} algorithm.
In this algorithm, the error in the classification obtained at the output layer is propagated to the previous layers using gradient descent~\cite{goodfellow2016deep}.
Each individual neuron then uses this error signal to update the weights associated with the output of the neurons in the previous layer.
These updates, performed multiple times over a set of samples provided in batches, enable the network to learn the hidden function that associates the inputs to their corresponding classes/outputs.

\section{Reducing Noise through Abstractions}\label{sec:prop}
To address the problems produced by the noise present in the data, we propose the use of data abstractions.
The aim of this proposal is that the simplification made by the abstraction would mitigate the influence of the noise present in the data.
This would be achieved through smoothing out the noise (and outliers) by abstracting them, together with other nearby data points, into a discrete space with fewer elements.

We define an abstraction as a mapping of the dataset values (origin set) to a finite set of values with a smaller cardinality (target set).
\begin{definition}
Given $A$ and $B$ two sets of numerical values, with $|A| > |B|$. We define an abstraction as a function $\mathcal{F}:A \to B$ that maps the values of $A$ to the values of $B$.

Given $X$, the bag of values to abstract, with $X \in A$, the abstraction of $X$ to $B$ is done by dividing $A$ into $n$ continuous subsets based on $n-1$ specific values that we call \emph{cut-off values}, with $n = |B|$, and mapping those subsets to the values of $B$. Then, each value in $X$ is assigned a value in $B$ based on its value in $A$.
\end{definition}


The cut-off values defined previously can be set using different approaches, and our first experiments focus on exploring which approach generates abstractions that get closer (if not better) accuracy to that obtained using the original raw data. 
To that end, we explore the use of static binning, quantiles, ROC Curves and K-Means clustering.

Notice that, independently of the chosen alternative, all the abstractions are performed on a ``per-column basis'' in the dataset, that is, they are performed per feature.
This is a crucial point, since using the same abstraction for different features will produce undesired effects, especially if different features have different orders of magnitude.

It is important to note that the alternatives explored in this section are a first approach to generate meaningful abstractions.
Further research into developing better abstraction generation approaches would be required, this paper being a proof of concept aimed at exploring the usefulness of research into abstracted data.
Thus, we focus our proposal on the benefit of abstractions, independently of how they were created.


\subsection{Static Binning}
Static methods to discretise data are common in statistical usage, specifically when studying the frequency or probabilities of continuous valued features~\cite{cetinkaya2019open}.
The most common approach is to divide the range of data $L$ into $n$ equal intervals of length $L/n$, where $L$ is the range of values in $A$, i.e., $L=\max{A}-\min{A}$.

\subsection{Quantiles}
Statistically, quantiles define cut-off points dividing a probability distribution into a fixed number of partitions.
Cut-off values are chosen in such a way that the probability distribution of the random variable ranging over the set $X$ is divide into intervals with equal probability $1/n$.
Formally, the data is divided according to the locations $y_q$ of the quantile $q$ ($q=k/n$, where $k\leq n$) given by 
\begin{equation}
 q = \int_{-\infty}^{y_{q}} p(x) dx.
 \label{eq:quantile}
\end{equation}
Quantile-based transformation results in elements of $B$ being uniformly distributed, therefore preserving the largest possible amount information for given $n$. 
Common quantiles include median ($n=2$), quartiles ($n=4$) and deciles ($n=10$).

\subsection{Receiver Operating Characteristic Curves}
Receiver Operating Characteristic (ROC) curves are a statistical methodology to analyse binary classifiers.
In our case, we will use the underlying concept to find the division of a feature into two classes that maximises balanced accuracy (given by Eq.~(\ref{eqn:BA})). 
Such a choice guarantees the best separation between two classes one wants to discriminate between.
ROC Curves are commonly used in healthcare modelling~\cite{metz1978basic} and as a metric for ML tasks which involve binary classification.

\subsection{K-Means Clustering}

K-means clustering~\cite{lloyd1982least} is a popular unsupervised data mining method.
It is mostly applied for grouping data points together, based on the similarities of their features, thus creating $k$ distinct clusters.
Abstractions are created such that elements of each cluster are mapped into unique bins, similar to automatic labelling of data.
The simplest implementation of K-means utilizes distance metrics to determine similarity

\section{Experiments}\label{sec:exps}
In this section we present the experiments we performed to validate our approach. Specifically, we performed four different experiments:
\begin{itemize}
    \item Comparing different approaches to generate abstractions.
    \item Measuring changes produced by noise in our abstractions.
    \item Comparing the performance of different ML methods using the raw data \textit{vs} using abstractions.
    \item Comparing the performance variance of different ML methods when faced with different noise scenarios, for both the raw data and the abstracted data.
\end{itemize}

To perform these experiments, we used six datasets of small to medium size.
These datasets, with binary classification as the goal, were chosen considering the volume of data and number of numerical features.
The characteristics of the datasets are displayed in Table~\ref{tab:exp_sub}.

\begin{table}[ht]
\caption{Characteristics of the experimental subjects.}
\begin{center}\scalebox{.9}{$
\begin{array}{c | c c c}
\hline
\textbf{Subject} & \textbf{Size} & \textbf{\# features} & \textbf{\# entries}\\
\hline
\multicolumn{4}{c}{\textbf{small size}}\\
\hline
\textbf{Ionosphere~\cite{data, Sigillito1989Ionosphere}} & 11,934 & 34 & 351\\
\textbf{Sonar~\cite{data, sonar}} & 12,480 & 60 & 208\\
\textbf{Wisconsin Breast Cancer~\cite{data, Wolberg1995Breast}} & 17,070 & 30 & 569\\
\hline
\multicolumn{4}{c}{\textbf{medium size}}\\
\hline
\textbf{Accelerometer~\cite{data, Scalabrini2019Prediction}} & 408,000 & 4 & 102,000\\
\textbf{Higgs~\cite{data, baldi2014searching}} & 2,800,000 & 28 & 100,000\\
\textbf{Air Pressure System Failure~\cite{data}} & 10,200,000 & 170 & 60,000\\
\hline
\end{array}$
}
\end{center}
\label{tab:exp_sub}
\end{table}

An important fact about these datasets is that all are clean curated datasets, except for the Air Pressure System Failure one, which contains empty values.
Thus, when performing our experiments, we will use this incomplete dataset only in the first experiment, while for the other experiments we will stick to the clean and curated datasets.
We took this decision mainly due to our extensive use of ANNs in such experiments, and thus, the lower relevance of exploring incomplete datasets.

We developed our experiments on Python~\cite{van2009python}, using scikit-learn~\cite{scikit-learn} for Logistic Regression, Random Forest and SVM methods, and tensorflow-keras~\cite{tf, chollet2015keras} framework to develop the ANN.
We ran our experiments using an Ubuntu server equipped with an NVIDIA A100 GPU with $80$GB of vram, $16$GB of ram and an Intel Xeon processor running at $2.20$GHz.

\subsection{Comparing abstraction generation processes}
Our first experiment is directed towards comparison of different approaches to abstraction generation.
For this purpose, we developed a pipeline for generating data abstractions using the different methods discussed in Section~\ref{sec:prop}.
Then, we trained the multiple ML methods discussed in Section~\ref{sec:prel} separately on the generated abstractions and on the raw data and compared the resulting accuracies.
To be precise, we explored generating abstractions using static binning into $2, 4, 10$ and $100$ bins, quantiles into $2, 4$ and $10$ quantiles, ROC Curves into $2$ bins, and using K-Means clustering into $2, 4, 10$ and $100$ clusters.
Finally, to reiterate, we used Logistic Regression, Random Forest, SVM, and ANN as ML methods.

We developed a small ANN to process our data, the architecture is as follows:
\begin{itemize}
    \item Input layer with as many neurons as input features.
    \item Hidden layer of size $100$ and LeakyReLU as activation function.
    \item Hidden layer of size $50$ and LeakyReLU as activation function.
    \item Hidden layer of size $25$ and LeakyReLU as activation function.
    \item Output layer of size $1$ and sigmoid as activation function.
\end{itemize}
We trained it with binary crossentropy loss and rmsprop optimiser~\cite{goodfellow2016deep}.

We define this basic architecture which works fairly well on our datasets to compare its performance when using two kinds of data: raw data and abstracted data.
Therefore, we do not foresee a need to search for a perfect neural network through hyperparameter optimisation.
In fact, we surmise that a very good neural network could be counterproductive, because it would generate very good results independent of the data.
Therefore, an ``average'' neural network is better for us to be able to properly compare its performance over our datasets.
As such, we utilise the same network for all our experiments.

\begin{figure*}[ht]
    \centering
    \includegraphics[width=0.45\textwidth]{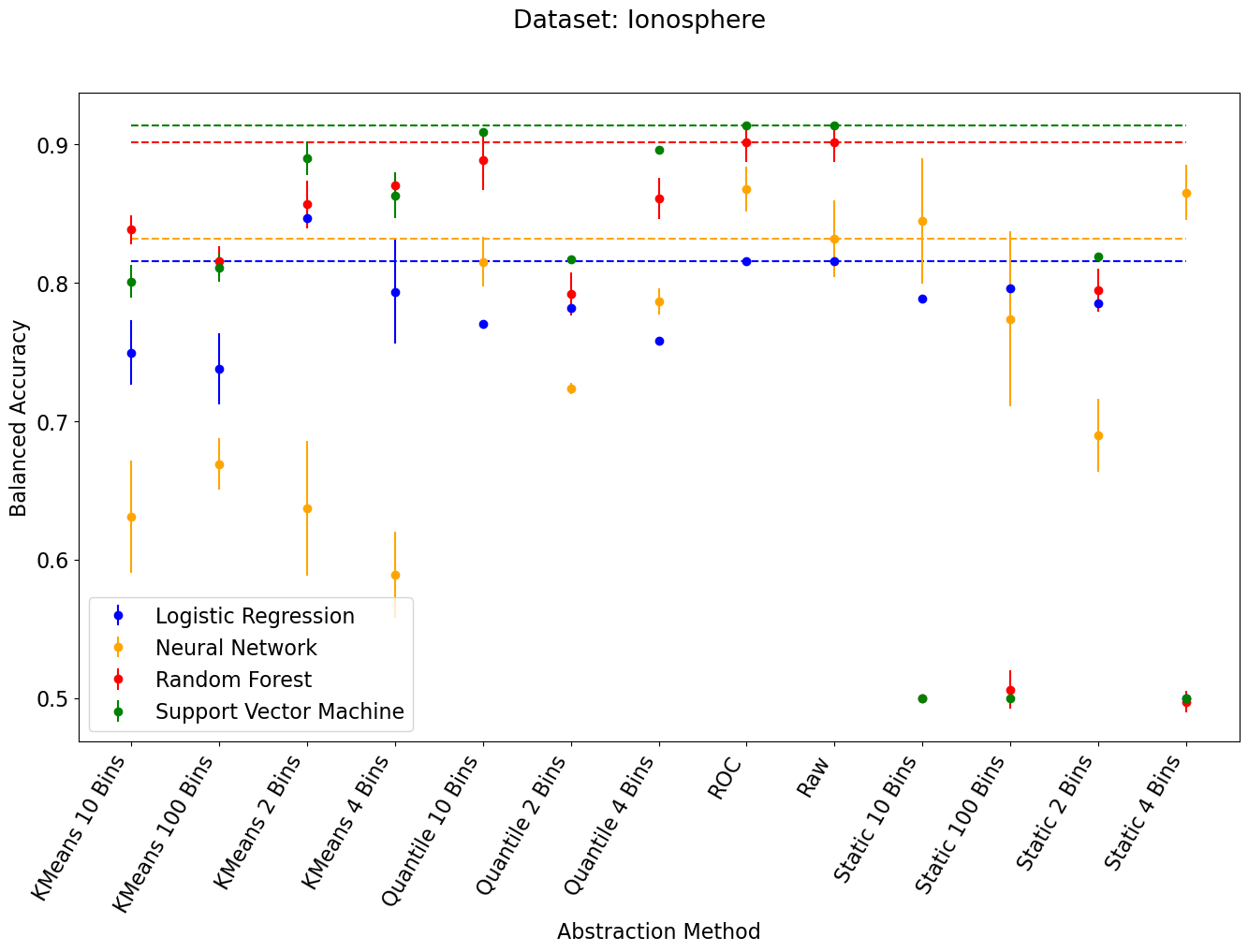}
    \hspace{1em}
    \includegraphics[width=0.45\textwidth]{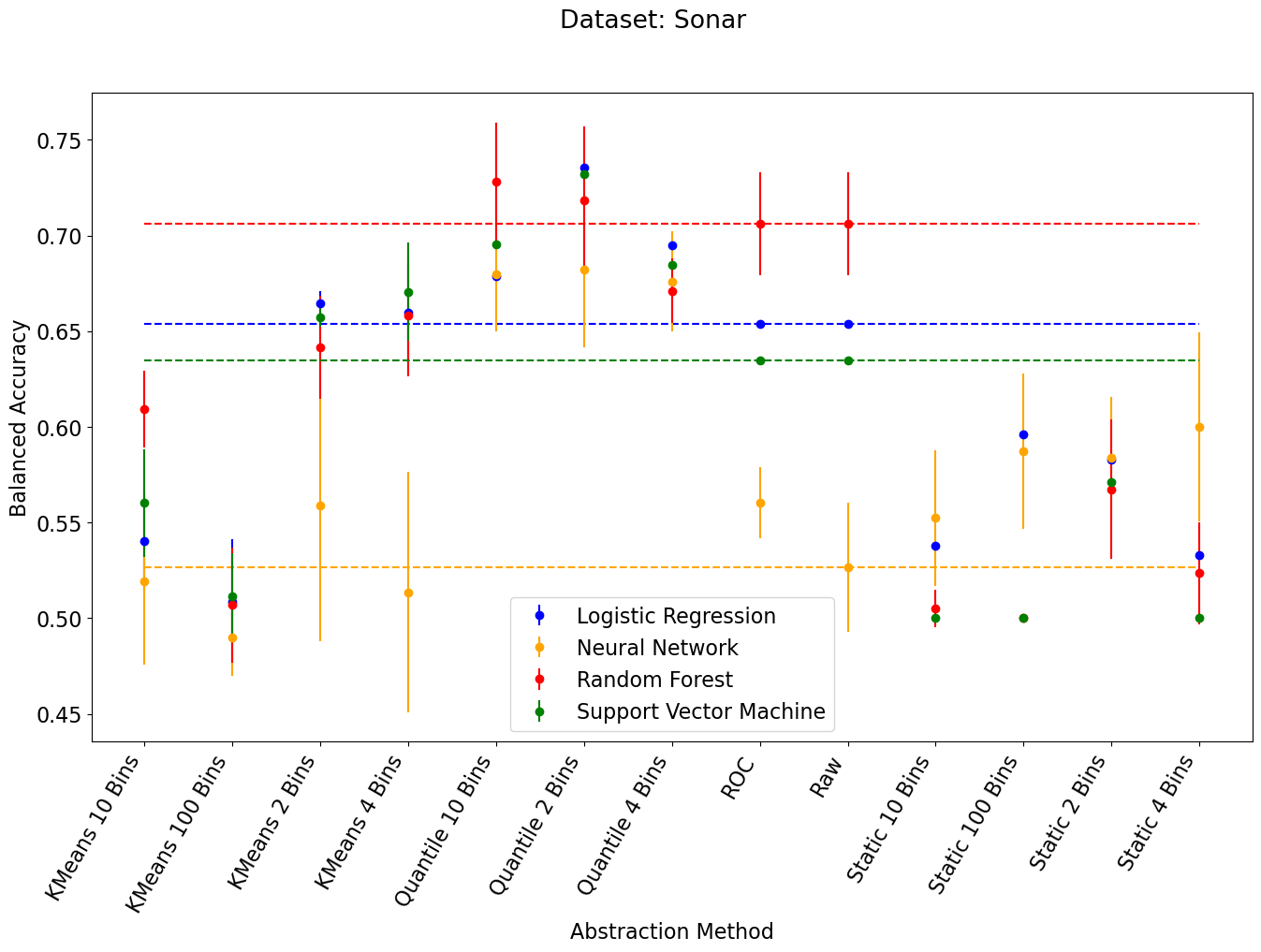}\\
    \vspace{1em}
    \includegraphics[width=0.45\textwidth]{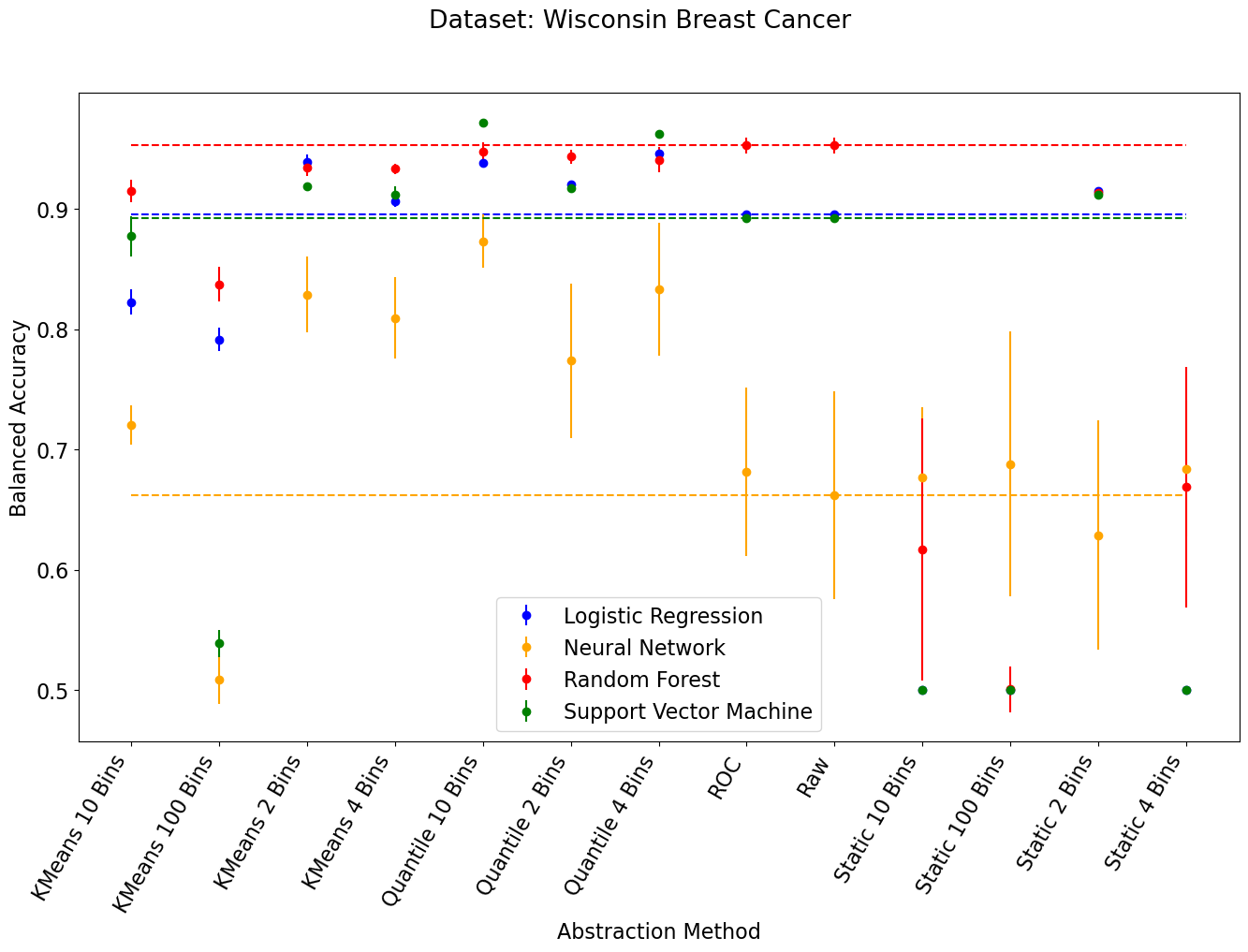}
    \hspace{1em}
    \includegraphics[width=0.45\textwidth]{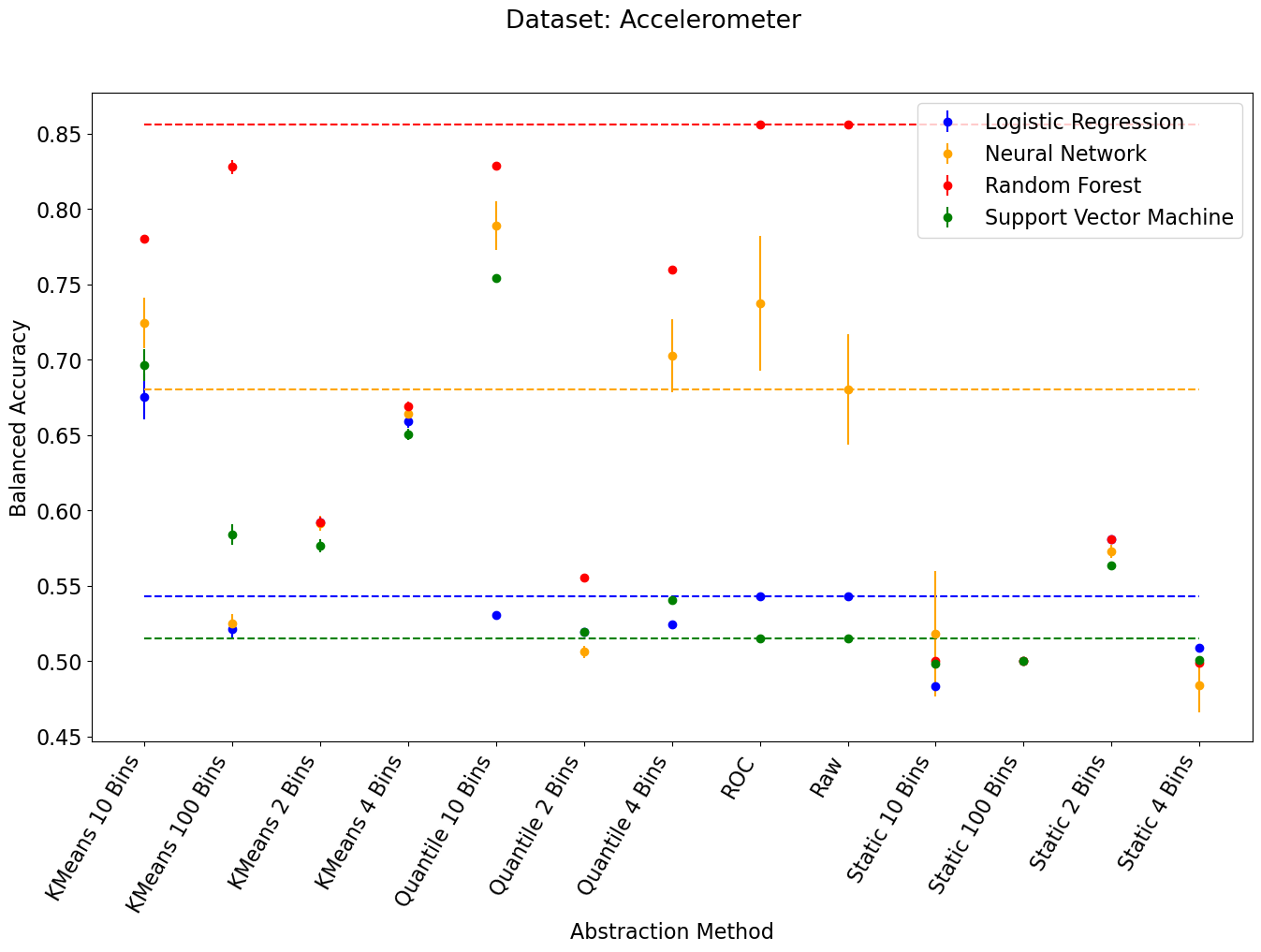}\\
    \vspace{1em}
    \includegraphics[width=0.45\textwidth]{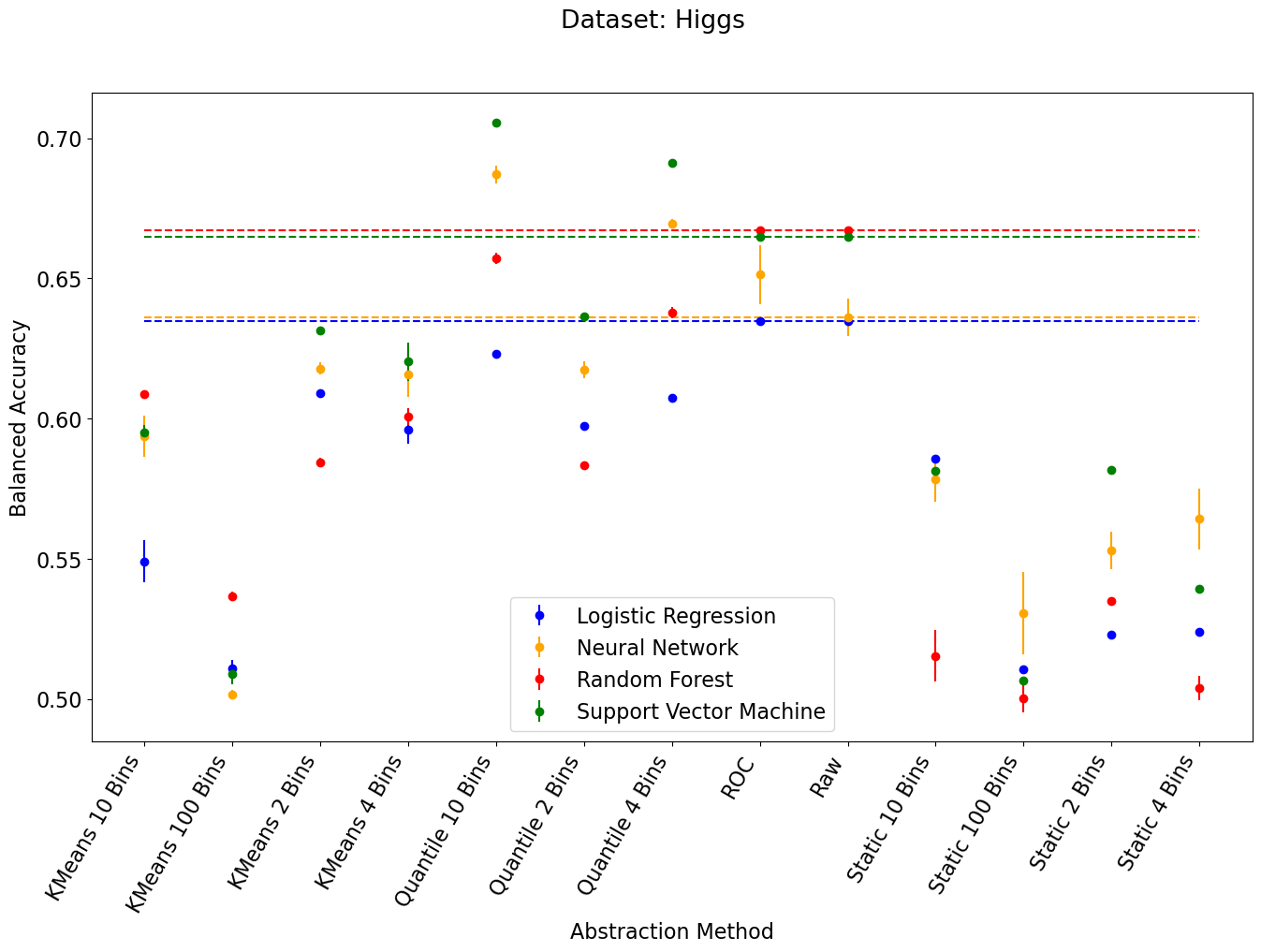}
    \hspace{1em}
    \includegraphics[width=0.45\textwidth]{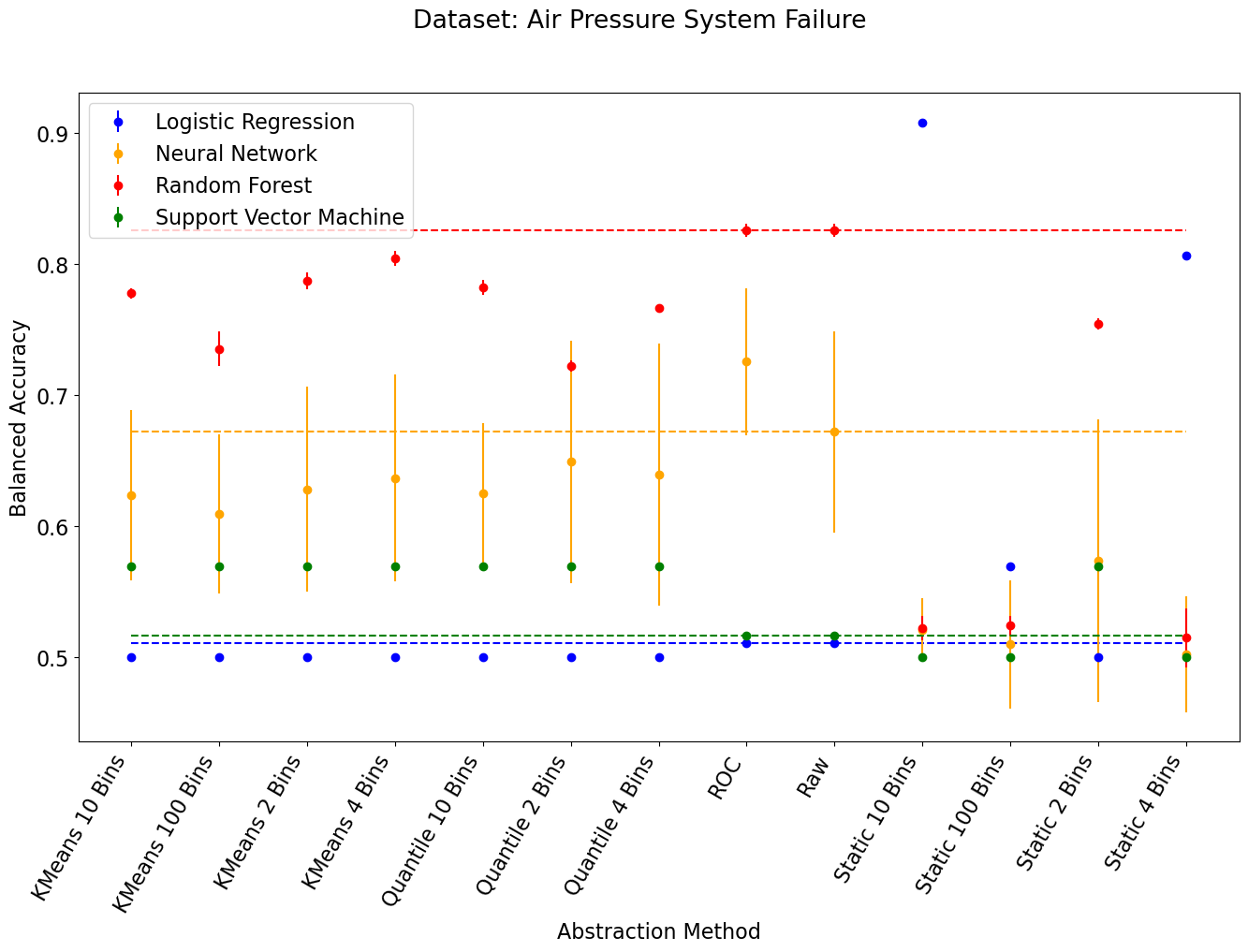}
    \caption{Results of the first experiment per dataset. Horizontal lines mark the raw data results for comparison.}
    \label{fig:results_1}
\end{figure*}

With this setup, and knowing the relevance that the random seed could have in our results, we repeated the experiments with $5$ different seeds to ensure the results were not product of a lucky random seed, fixing the same seed for all the methods each time.
The results of our experiments are presented in Figure~\ref{fig:results_1}.

An analysis of these plots shows that better results (on average) correspond to the use of ROC Curves to abstract data.
However, there are a bit of nuance in these results.
Specifically, abstracting using ROC Curves do not always arise the best accuracy, as is the case for the Sonar, Wisconsin Breast Cancer, and Higgs datasets.
In those cases, abstracting into $10$ quantiles is a better option, as it improves accuracy, and specifically for Wisconsin Breast Cancer and Higgs datasets it is the best option. However, for the Sonar dataset abstracting into $2$ quantiles is the best option.
The results of our experiments show us that deciding a method of abstraction for a dataset is not an easy task since it strongly depends on the features, dataset and the ML method used.

Additionally, it is important to remark that a curious effect is noticeable using abstractions in certain scenarios: they get higher accuracy than raw data.
For certain tasks, it could be better to use a meaningful abstraction that comprises the relevant information, than using the raw data which could have potentially irrelevant, maybe even misleading, information.
This can be observed in the results for the Sonar dataset, where ROC Curves and the quantile alternatives get higher accuracy, with all the ML methods, than using raw data.

The clear conclusion of these experiments is that ROC Curves are a suitable method for generating abstractions. Except in the Sonar, Wisconsin Breast Cancer and Higgs datasets where $10$ bin quantiles (deciles) are the best option.
Thus, in the rest of our experiments we use ROC Curves and deciles to generate the required abstractions.

\subsection{Abstraction and robustness to noise}
\begin{figure*}[ht]
    \centering
    \includegraphics[width=\textwidth]{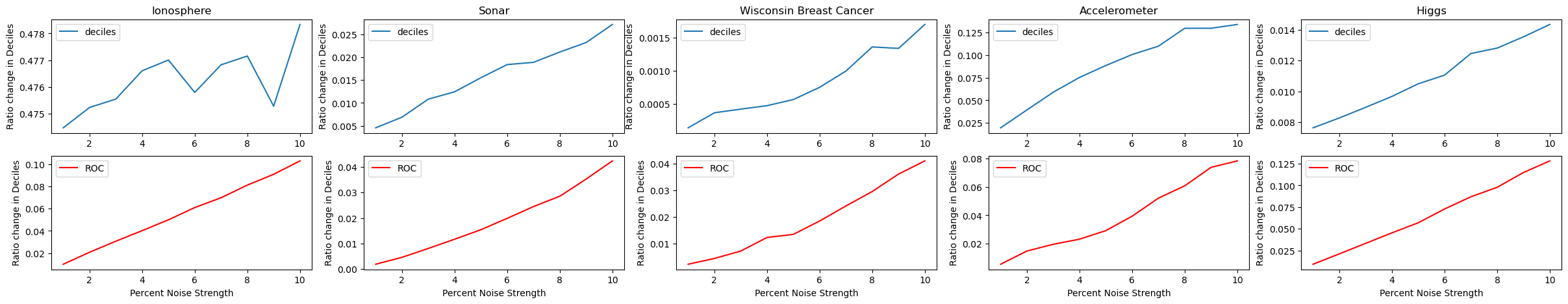}
    \caption{Mean of abstraction boundary variance for different noise strengths.}
    \label{fig:results_2}
\end{figure*}

In this experiment we aim to explore how much an abstraction can change when computed over clean data vs when computed over noisy data.
To measure this we considered the percentage change of classes between the different noise levels.
To perform this experiment we introduced Gaussian noise in all values of the dataset, with a strength of up to $10\%$ of the data range of the feature.
We used this approach following classical literature in adding noise to data~\cite{silva2011pca}, with the goal of representing Gaussian random noise.
Then, we computed two abstractions, one using ROC Curves and another into deciles, as explained in Section~\ref{sec:prop}.
Additionally, we computed the corresponding abstractions for the clean dataset and then we compare both abstractions, measuring the percentual change of bin size. 
As noise is introduced randomly, we repeated the experiment with $5$ different seeds and averaged over them.
The results of these experiments are displayed in Figure~\ref{fig:results_2}, where we show the average percentual change (in absolute value) for each abstraction, for each dataset.

Analysing our results, we observe how noise could produce a huge change in the abstraction bins, with a maximum of $40\%$ variance. 
However, if we choose the proper method, we can achieve a maximum variance of $10\%$.
Moreover, we can observe how there is no direct relation between the variance and the number of bins.
Also, it is clear that, depending on the dataset, a different alternative for abstracting data is preferable. 
This can be observed by the fact that, depending on the dataset, a different alternative is the one having less variance.
Thus, further work should analyse which criteria is fundamental to decide which abstraction will be less affected by noise for a given dataset.

\subsection{Comparing ANN performance}
For this experiment we took the same ANN that we used before as the method of reference to explore its robustness to noise. 
With it, we trained three models: one with the raw data, one with the data abstracted using ROC Curves, and one with the data abstracted into deciles. 
For validating the outputs of this task, we divided the data into two sets \textit{apriori}: a training set and a test set, and we used the same division for all the models, as is common in ML pipelines.
We additionally performed standard MinMax Scaling~\cite{scikit-learn} for the raw data to reduce all values in each feature to the range $[0, 1]$.
Then, we trained all models with their respective training sets and evaluated their performance with their respective test sets. We repeated the whole experiment $10$ times and reported the averaged results.

We repeated this whole experiment with the original noise-free data and with noisy data. 
For our experiments, we added noise in three ways: adding noise to the training and test datasets, adding noise only to the training dataset, and adding noise only in the test dataset. 
The goal of adding noise in both datasets is to validate how much noise changes training and testing, while the goal of adding noise only in the training dataset is to validate how much a noisy training affects the overall performance. 
Finally, the goal of adding noise only to the test dataset is to explore how abstractions can mitigate the problem of having noisy data as inputs after training.

We introduced Gaussian noise in all values of the dataset, with a strength of up to $10\%$ of the data range of the feature. 
The goal is to explore the worst case scenario where noise is present across the dataset.
Finally, it is important to note that, in order to have control over the pseudo-randomisation process, we set a global random seed. This seed is then set to the machine's time in order to introduce the noise in the dataset randomly, but afterwards it is set again to the global seed. 
This implies that the initial parameters of the different methods are always the same, which allows us to properly compare the obtained accuracy values.

\begin{figure*}[ht]
\centering
    Clean Training and Testing\\
    \vspace{0.25em}
    \includegraphics[width=1\textwidth]{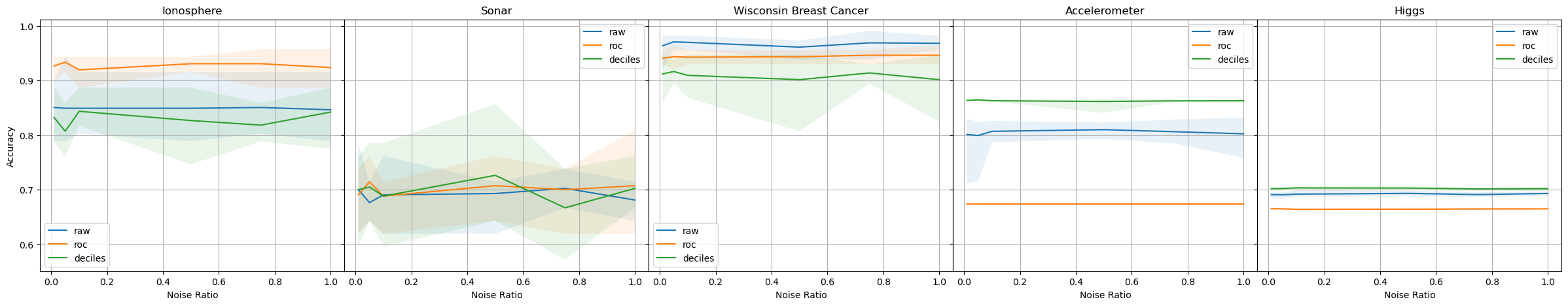}\\
    \vspace{1em}
    Noisy Training\\
    \vspace{0.25em}
    \includegraphics[width=1\textwidth]{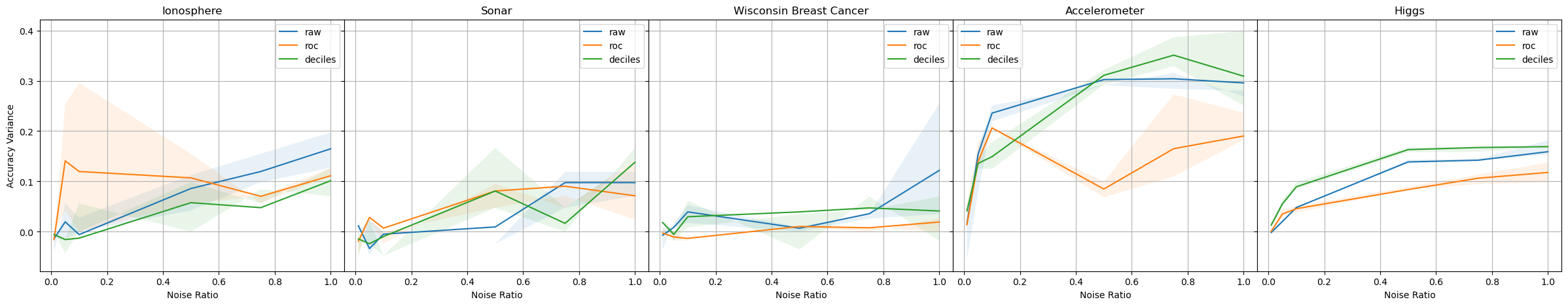}\\
    \vspace{1em}
    Noisy Testing\\
    \vspace{0.25em}
    \includegraphics[width=1\textwidth]{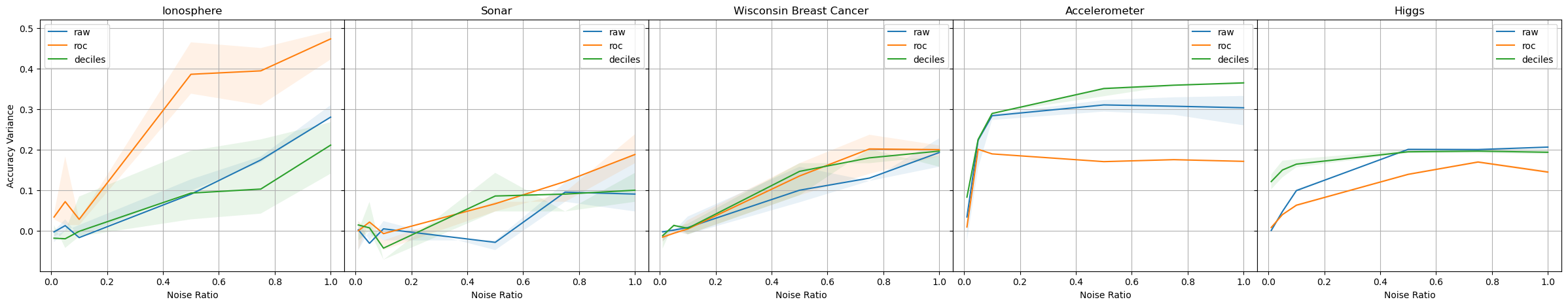}\\
    \vspace{1em}
    Noisy Training and Testing\\
    \vspace{0.25em}
    \includegraphics[width=1\textwidth]{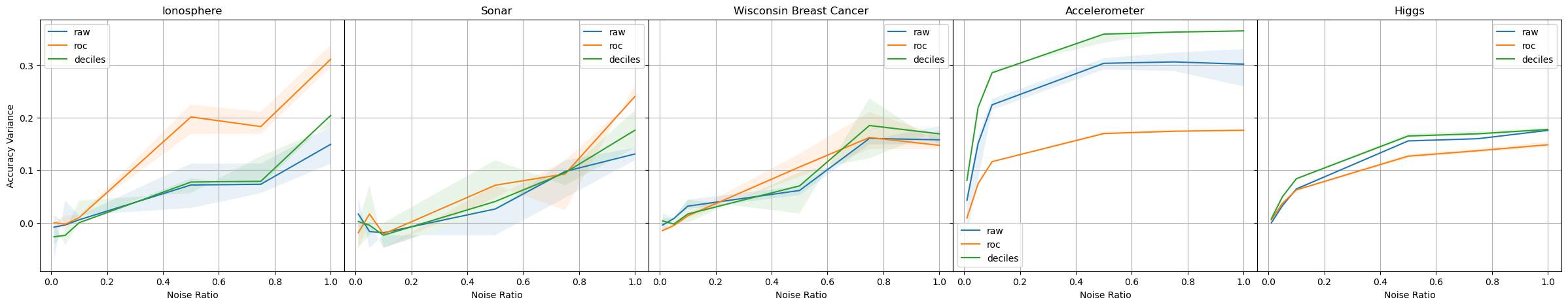}
\caption{Variation of accuracy scores per noise scenario, for the different datasets, and per abstraction, with respect to Clean Training and Testing (scenario where we show the actual accuracy scores obtained).}
\label{fig:vari_acc}
\end{figure*}


\subsubsection{Clean training and testing}

We started with the ideal case, where we used noise-free datasets both during training and runtime. 
For example, this would be the case when dealing with medical laboratory data, that has followed multiple quality and integrity checkups to ensure its reliability. 
Then, our hypothesis was that raw data would perform better than the abstracted data, although we expect the difference to be small.

The experiment resulted in the accuracy values displayed in Figure~\ref{fig:vari_acc} in the first row. 
There, we can observe how, sometimes, the abstracted data (either using deciles or ROC Curves) gets higher accuracy than the raw data. 
This is a surprising result we were not expecting to achieve, but that reinforces the idea that abstracting the data does not always imply a loss in accuracy. 
Thus, these results show that using abstractions do not hamper the classification efforts of a neural network. 
In fact, sometimes even reinforces such efforts.

An interesting result here is that the abstracted data results in the ANN getting higher accuracy and loss scores in training than the raw data. However, we also obtained an overfitting effect, getting lower loss (and higher accuracy) in training but higher loss (and lower accuracy) in test, which is not ideal. We consider that this shows how abstractions can improve the learning phase of an ANN. However, this could not be desirable in every situation due to the risk of overfitting. In any case, it is matter of future work to explore how to properly use this improvement in the learning phase to obtain better scores in testing, what could potentially imply that learning using abstractions is better than using raw data.

\subsubsection{Noisy training and clean testing}
When adding noise only to the training dataset, the idea is to check how much noisy data can hamper the learning efforts of a neural network when the noise is not reproduced at runtime. 
For example, this would be the case when you have trained your model with measures from a sensor with high error margins, and later you apply the model over data from an updated sensor with lower error margins. 
In this case, our hypothesis was that the abstracted data would be less affected by noise, and therefore would be able to better generalise from such noise.

After executing the experiment, we got the variation of accuracy scores displayed in Figure~\ref{fig:vari_acc} in the second row. There, we can observe how, in general, the smaller variation corresponds to the abstracted data, while the raw data tends to have higher variations. 
However, as always, the results depend on the dataset, with data abstracted using ROC Curves being better for the big datasets (Higgs and Accelerometer) while deciles is better for the smaller datasets (Ionosphere and Sonar).

In this case, we observe that the variation in accuracy is smaller for the abstracted data, although in the case of raw data there is a dataset where they are at par. 
Thus, we can conclude that the abstracted data is a better approach in this situation, because it is able to better generalise from noisy data. 
Therefore, we can conclude that this hypothesis was correct too.

\subsubsection{Clean training and noisy testing}
When adding noise only to the test dataset, the idea is to check how much noisy data can hamper the classification efforts of a neural network when the noise is not present at learning time. 
For example, this would be the case when you have a very curated dataset at training but later at runtime the data is not processed enough and it is more noisy than expected. 
This would be also the case of adversarial attacks. Our hypothesis in this case was that this noise influence will be slightly mitigated by the abstraction function and therefore it will affect less the neural network trained over the abstraction.

After executing the experiment, we got the variation of accuracy scores displayed in Figure~\ref{fig:vari_acc} in the third row. There, we can observe how, in general, the smaller variation corresponds to the abstracted data, while the raw data tends to have higher variations. 
In this case the data abstracted using ROC Curves is better for the big datasets (Higgs and Accelerometer) while deciles is better only for the Ionosphere dataset, having raw data the smallest variation for the Sonar and Wisconsin Breast Cancer datasets.

Once again, we can conclude that using abstractions is a better option, as the abstracted data obtains lower accuracy variations, but in this case raw data also obtained good results for two datasets. 
Thus, once again, we can conclude that our hypothesis was correct.

\subsubsection{Noisy training and testing}
When adding noise to both training and test dataset, the idea is to check how much noisy data can hamper the performance of a neural network, if that noise is replicated also at runtime. 
For example, this would be the case when you have measures from noisy sensors, that is, sensors with big error margins. 
Our hypothesis was that this noise will hamper more the neural network trained over the raw data than the neural network trained over the abstracted data.

After executing the experiment, we got the variation of accuracy scores displayed in Figure~\ref{fig:vari_acc} in the last row. There, we can observe how, in general, the smaller variation corresponds to the abstracted data, while the raw data tends to have higher variations. In this last case the data abstracted using ROC Curves is still better for the big datasets (Higgs and Accelerometer) while deciles do not manage to obtain smaller variations for any dataset, having raw data the smallest variation for the Ionosphere and Sonar datasets.

With these results we can safely conclude that the abstracted data was more robust in the presence of noise and was able to get a smaller variation in accuracy than the raw data, although the results are not uniform across datasets. Thus, we can conclude that our hypothesis was correct.

\subsection{Comparing noise effect on ANN performance}
After our first experiments, we wanted to check how much accuracy variation do the different noise scenarios introduce in the ANNs performance. To that end, we took the same ANN than in the previous experiments, and run it in the four different noise scenarios explored in the previous experiment. We also used the same random seed setup than in the previous experiments. In this case, we do not only record the accuracy and loss values for each scenario, but we also computed their variance along the four scenarios. We repeated this experiment $10$ times and present averaged results.

\begin{figure*}[ht]
\centering
    Average Accuracy\\
    \vspace{0.25em}
    \includegraphics[width=1\textwidth]{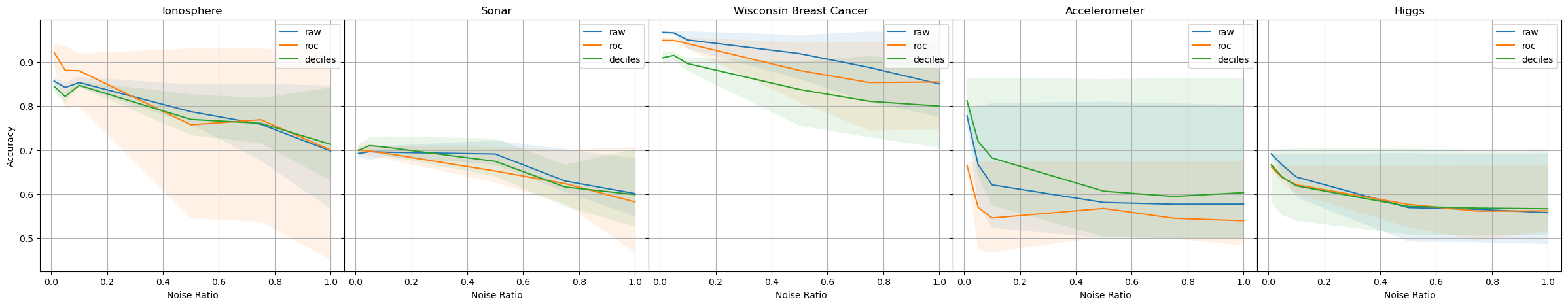}\\
    \vspace{1em}
    Standard Deviation\\
    \vspace{0.25em}
    \includegraphics[width=1\textwidth]{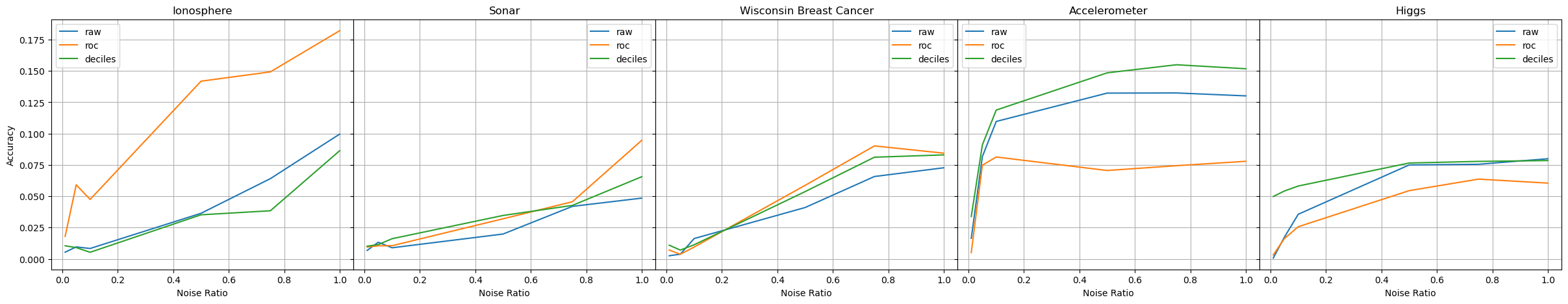}
\caption{Average accuracy (top) and standard deviation (bottom) between noise scenarios.}
\label{fig:avg_and_std}
\end{figure*}

In Figure~\ref{fig:vari_acc} we display in the top row the average accuracy for each dataset, for each different source data, for each noise level, along the maximum accuracy and minimum accuracy values. In those plots, the solid line is the average accuracy, and the fade-out area is the range between the minimum and the maximum accuracies. In the bottom row we present the standard deviation between noise scenarios for each dataset, for each different source data, for each noise level.

As we can clearly observe, the variation of accuracy is lower for the data abstracted into deciles when dealing with smaller datasets (Ionosphere and Sonar datasets) while the variation of accuracy is lower for the data abstracted with ROC Curves for bigger datasets (Accelerometer and Higgs). This implies that abstractions work better along the four scenarios we have explored than raw data, what confirms our hypothesis that abstracting data could have benefits under noisy scenarios.

\subsection{Experiments summary}
To summarise our results, we have shown with our experiments that abstracted data is more robust to noise than raw data, independent of the presence of the noise in the training and testing datasets, only in the training dataset, or only in the testing dataset.
This is an important result, as it validates our hypothesis that the use of data abstraction can help mitigate influence of noise to develop robust ML models.
However, not every abstraction is useful for every dataset, and our results suggest that using ROC Curves is better for bigger datasets while using deciles would be better for smaller datasets, although further work is needed to confirm this.

Additionally, we have discovered that using abstracted data, backpropagation models are able to better fit to the training data.
This result is worth considering since, translating this improvement to the test phase, abstractions could theoretically improve current state-of-the-art results, although further analysis of this phenomenon is necessary.

Finally, it is important to remark that the results of our experiments on the ANN are pertinent when considering that abstractions increase the number of data value changes introduced by the noise. In fact, on average we are increasing those changes a $10\%$.
However, as observed, those extra changes do not affect the overall performance of the ANN, as the variance in accuracy is lower when using the abstracted data.
This is an interesting effect that could be analogous to the abstraction smoothing out noise between multiple values instead of having it centred in only one value.



\section{Threats to Validity}\label{sec:tstv}
In this section we discuss the possible threats to the validity of our results.
The first kind of threats we consider are the threats to internal validity, that can explain our results due to uncontrolled factors.
The main threat in this category is the possibility of having a faulty code.
To reduce this threat we have carefully tested each piece of code used in our experiments and we have relied on widely tested libraries like tensorflow-keras.
Another threat in this category is the impact of randomisation in our results.
To control this factor we have repeated our experiments several times and averaged results.
We have also controlled the random seeds when needed, for both reproducibility and comparison purposes.

The second kind of threats are the threats to external validity that hamper the generality of our results to other scenarios.
In our case the only threat in this category is the small scale experimental setup, having tested four ML methods over six datasets.
However, we have performed small experiments with other ML methods and datasets too, and obtained similar results.
Nonetheless, the exploration of the robustness of abstractions to noise for a broader set of ML methods remains part of future work.

Finally, the last kind of threats are the threats to construction validity, hampering the extrapolation of our results to real-world scenarios.
In this case, the range of possible scenarios is potentially infinite, and this threat cannot be fully addressed, but as explained before, the exploration of the robustness of abstractions to noise for other scenarios is matter of future work.

\section{Conclusions}\label{sec:conc}
Noise robustness is a desirable property in any Machine Learning (ML) application, as it is usually difficult to obtain clean, curated, real world data.
When applying ML methods to real-world datasets that have not been curated by a human, having methods that can overcome the potential noise is a fundamental need.

To address this problem, we proposed a solution in this paper: using data abstractions to mitigate noise influence.
To that end, we have developed two abstraction mechanism based on ROC Curves and quantiles, that try to generate the best abstraction for the data at hand. These abstractions are computed over the training dataset, and later applied over any new data the ML method encounters.

To evaluate our proposal, we performed several experiments, exploring how noise affects a basic ANN performance when introduced in both training and testing datasets, only in training dataset and only in testing dataset.
From these experiments, we conclude that using abstractions is better in situations involving noisy data, due to the capability of mitigating the influence of noise.
Moreover, we discovered that with our approach, using abstractions improves the training of the neural network, although this improvement does not reflect in the testing phase.
Finally, we also discover that abstracting using ROC Curves tends to be a better option for bigger datasets than abstracting to deciles, that tends to be better for small datasets.

In future work, we would like to explore the effect of using abstractions with other ML methods and over other datasets, to develop a way of extending the improvement in the training phase to the testing phase.
Additionally, we would like to explore how abstractions could reduce the effects of adversarial attacks, as it is theoretically possible to mitigate  the adversarial attack mask effect.
We would also like to explore the noise reduction effect over other datasets and models.
We would like to explore the effect of binary abstractions over non-binary classification task and the impact of abstractions on bias. 
We would also like to extend our approach to deal with categorical data, text data, and/or temporal data.
Finally, we would like to explore how to properly apply the abstractions to new data while updating the abstractions.

\section*{Acknowledgments}
This research has been supported by the European Union’s Horizon $2020$ research and innovation programme under grant agreement Sano No $857533$.
This publication is supported by Sano project carried out within the International Research Agendas programme of the Foundation for Polish Science, co-financed by the European Union under the European Regional Development Fund.
This research was supported in part by PLGrid Infrastructure.

\bibliographystyle{plain}
\bibliography{biblio}

\newpage

\vfill

\end{document}